\documentclass[runningheads]{llncs}
\usepackage[T1]{fontenc}
\usepackage{graphicx}
\usepackage{amsmath,amssymb,amsfonts}
\usepackage{algorithmic}
\usepackage{comment}
\usepackage[flushleft]{threeparttable}
\usepackage{xspace}
\usepackage{graphicx}
\usepackage{textcomp}
\usepackage[table]{xcolor}
\usepackage{hyperref}
\usepackage{subcaption}
\usepackage{orcidlink}

\newcommand{\linf}{\ensuremath{L_{\infty}}\xspace}

\newcommand{\pointset}{\ensuremath{P}\xspace}
\newcommand{\nbpoint}{\ensuremath{n}\xspace}

\newcommand{\lebesgue}{\ensuremath{\lambda}}
\newcommand{\lebesguebox}[1]{\ensuremath{\lebesgue(#1)}}

\newcommand{\linfstardisc}{\ensuremath{d^*_{\infty}}\xspace}
\newcommand{\linfstardiscpointset}[1]{\ensuremath{\linfstardisc(#1)}\xspace}

\newcommand{\p}{\ensuremath{x}}
\newcommand{\point}[1]{\ensuremath{\p^{(#1)}\xspace}}

\newcommand{\fibo}{\ensuremath{F}}
\newcommand{\fibopointset}[1]{\ensuremath{\fibo_{#1}}\xspace}

\newcommand{\kronecker}{\ensuremath{K}}
\newcommand{\kroneckerpointset}[1]{\ensuremath{\kronecker_{#1}}\xspace}
\newcommand{\shiftedkroneckerpointset}[1]{\ensuremath{^{+1}\kronecker_{#1}}\xspace}

\newcommand{\irace}{\textsc{Irace}\xspace}

\newcommand{\parameterTwo}{\ensuremath{p_2}\xspace}
\newcommand{\parameterThree}{\ensuremath{p_3}\xspace}
\newcommand{\hide}[1]{}

\begin{document}
\title{Generating Point Sets of Low Star Discrepancy by Optimizing Kronecker Constructions}
\titlerunning{Low-discrepancy Point Sets in $3D$ Using Kronecker Constructions}
%
\author{Imène Ait Abderrahim  \inst{1,2}\orcidlink{0000-0002-5654-2058} \and
Carola Doerr\inst{1}\orcidlink{0000-0002-4981-3227} \and
Martin Durand\inst{1}\orcidlink{0000-0001-9117-8661}}
\authorrunning{I. Ait Abderrahim et al.}

\institute{Sorbonne University, CNRS, LIP6, F-75005 Paris, France \and
Khemis Miliana University, Ain Defla, Algeria}
%

\maketitle              
\begin{abstract}
The \linf star discrepancy of a point set is a measure for how uniformly a point set is distributed in a given space. Point sets of low star discrepancy are used as designs of experiments, as initial designs for Bayesian optimization algorithms, for quasi-Monte Carlo integration methods, and many other applications. Recent work has shown that classical constructions such as Sobol', Halton, or Hammersley sequences can be outperformed by large margins when considering point sets of fixed sizes rather than their asymptotic behavior when the number of points tends to infinity. These results, highly relevant to the aforementioned applications, raise the question of how much existing constructions can be improved through size-specific optimization. 
In this work, we study this question for the 3-dimensional setting, focusing on how well the so-called Kronecker construction performs for various set sizes $n$. We show that, for settings with at least 500 points, optimizing the two configurable parameters of its construction using evolutionary computation approaches such as CMA-ES yields point sets whose discrepancy values outperform those of the previously best-known sets. Using the algorithm configuration technique \irace, we then derive parameters that yield new state-of-the-art discrepancy values for whole ranges of set sizes. An investigation of the problem landscapes using exploratory landscape analysis reveals that the problem is highly multimodal and that some of its characteristics differ substantially from the 24 BBOB functions. 

\keywords{Low-discrepancy point sets  \and uniformity of distribution \and Evolutionary computation}
\end{abstract}

\section{Introduction}
Discrepancy measures are designed to quantify how well-distributed a set of points is in a given space. Among the various discrepancy measures, the \linf star discrepancy is one of the most important. 
It measures how well spread the points in a point set P are. It is defined as the maximum absolute difference between the volume of a box anchored in the origin and the proportion of points from the P falling inside it. 
Indeed, the Koksma-Hlawka inequalities~\cite{koksma1942general,hlawka1961funktionen} show that it is possible to bound the error made on the approximation of an integral by the average of a finite number of local evaluations: The smaller the discrepancy of the point set, the smaller the error guarantee for the quasi-Monte Carlo approximation. Although its primary use is in quasi-Monte Carlo integration \cite{dick2010digital}, point sets with minimal \linf star discrepancy are also employed in one-shot optimization \cite{cauwet2020fully}, experiment design \cite{santner2003design}, computer vision \cite{paulin2022matbuilder}, and financial mathematics \cite{galanti1997low}.
Optimizing \linf star discrepancy is a computationally challenging problem. Indeed, optimal point sets are known only for up to 21 points in dimension 2 and for up to 8 points in dimension 3~\cite{clement2025constructing}.   
For the 2-dimensional hypercube, the Fibonacci set gives consistently good point sets, regardless of the size~\cite{clement2025constructing}. It builds the coordinates in one dimension using the golden ratio and in the other using the inverse of the number of points. This ensures that the points are spread very uniformly in both dimensions. No such method exists for higher dimensions. The construction of low \linf star discrepancy point sets has traditionally relied on number-theoretic sequences such as Halton~\cite{Halton60}, Hammersley~\cite{Ham60}, or Sobol’~\cite{Sobol}, which offer strong asymptotic guarantees but were recently shown to be less suitable for sets of fixed size~\cite{clement2024heuristic}. 
To address this limitation, research on the optimization-driven construction of low-star-discrepancy point sets tailored to specific dimensions and sample sizes has received increased attention in recent years.

\subsection{Related work}
To obtain point sets of low discrepancy, a broad range of different techniques have been used: exact linear programming techniques (in isolation~\cite{clement2025constructing} or as components of heuristic search~\cite{clement2025searching}), graph neural networks~\cite{rusch2024message}, 
large language models~\cite{sadikov2026generation}, heuristic search approaches~\cite{clement2024heuristic,clement2025low,clement2023computing,doerr2013constructing}. 

More precisely, one of the first approaches to tackle the optimization of low discrepancy is the work of Doerr et al. \cite{doerr2013constructing}, who propose a genetic algorithm that optimizes the generation of points from generalized Halton sequences to directly produce low-star-discrepancy point sets. Their hybrid evolutionary framework integrates efficient discrepancy estimation with stochastic search. In addition, Patel and Education \cite{patel2022optimizing} revisit designs based on lattices and sequences using parameter optimization instead of structural redesign. Specifically, optimized Kronecker sets show that uniformity can be significantly increased while maintaining simplicity, scalability, and progressive sampling features through careful empirical selection of irrational parameters. Another approach of Clément et al. \cite{clement2024heuristic,clement2022star}  investigates exact and heuristic strategies for selecting subsets of points that minimize star discrepancy. The study formulates the construction problem as a combinatorial optimization task and evaluates greedy, local search, and evolutionary approaches. 
    
In more recent work, point-set generation has been reframed as a geometric learning problem using learning-based techniques. In Rusch et al. \cite{rusch2024message}, they present Message-Passing Monte Carlo (MPMC), a graph neural network architecture that achieves contemporary performance for small to medium dimensions by learning point set transformations to minimize discrepancy-related objectives. Clément et al. \cite{clement2025searching} propose a permutation-based decomposition that separates relative point ordering from coordinate optimization, achieving state-of-the-art discrepancy reductions compared to classical sets. This decomposition yields a post-processing procedure, which we use in our experiments, to lower the discrepancy of a point set. Intuitively, this procedure maintains the relative position of points in all dimensions. It finds an optimal point set among the ones which preserve these relative positions. Clément et al.
\cite{clement2025low} propose a projected gradient descent algorithm for optimizing $L_2$ discrepancies in low dimensions. The main contribution is demonstrating that this computationally inexpensive method can achieve results comparable to or better than state-of-the-art techniques, even for $L_{\infty}$ star discrepancy, when starting with a reasonably uniform point set like the Fibonacci lattice.  
Sadikov \cite{sadikov2026generation} explores the emerging use of large language models (LLMs) as generative and guidance mechanisms for constructing low-discrepancy point sets.

\subsection{Contributions and outline of the paper}
In this paper, we investigate Kronecker sets for the construction of point sets of low star discrepancy. Kronecker constructions which generalize the 2-dimensional Fibonacci set to larger dimensions. Kronecker sets have $D-1$ degrees of freedom, i.e., parameters that need to be set to obtain a concrete set, one for each dimension. With the two-dimensional setting largely explored in previous works~\cite{rusch2024message,clement2025searching,sadikov2026generation}, we focus our work on constructing low-discrepancy point sets in dimension $3$. That is, we are dealing with a 2-dimensional optimization problem to identify the two parameters that yield good 3-dimensional point sets. 

We use evolutionary computational methods such as CMA-ES and local search variants to identify parameters that minimize the discrepancy of point sets of a fixed size, and we use the \irace configurator to obtain parameters that yield constructions for broad ranges of point set sizes. Even though this generation method is quite generic, it remains competitive with state-of-the-art solutions for small sizes and outperforms the best-known methods for larger sizes starting from 500 points.

We also investigate the landscape of the 2D-optimization problem using exploratory landscape analysis~\cite{mersmann2011exploratory_ela} and compare it to the well-studied BBOB functions from the COCO environment~\cite{Hansen02012021_coco}. 

We then briefly study the 4-dimensional Kronecker constructions, but we were unable to identify parameter values that would yield constructions outperforming the current state of the art, indicating that the Kronecker constructions may not be the right approach for dimension 4 (and possibly higher).

\textbf{Outline of the work.} 
In Section~\ref{sec:preliminaries}, we provide relevant background information such as the formal definitions of the \linf star discrepancy and the Kronecker sets. In Section~\ref{sec:algorithms}, we describe the methods used to tune the parameters of Kronecker sets. Our experimental results are detailed in Section~\ref{sec:experiments}. \ref{sec:landscapeAnalysis} provides an analysis of the problems landscape. Finally, we complete the paper with a discussion and a conclusion in Sections~\ref{sec:discussion} and~\ref{sec:conclusion}.

\textbf{Availability of code and data.} Our code and data are publicly available at~\hyperlink{https:/gitlab.lip6.fr/durandm/low_discrepancy}{https://gitlab.lip6.fr/durandm/low\_discrepancy}.
 
\section{Preliminaries}
\label{sec:preliminaries}

\subsection{\linf star discrepancy}

The \linf star discrepancy of a point set \pointset measures how well spread the points in \pointset are in the $d$-dimensional $[0,1)$ cube. It is defined as the maximum absolute difference between the volume of a box anchored in the origin and the proportion of points from \pointset falling inside it. Formally, given a point set \pointset of $\nbpoint$ points in $[0,1)^d$, the \linf star discrepancy of \pointset, denoted by \linfstardiscpointset{\pointset}, is defined as:

\begin{equation}
    \linfstardiscpointset{\pointset}= \sup\limits_{q \in [0,1)^d} \Bigg|\frac{D(q,\pointset)}{|\pointset|}-\lebesguebox{q}\Bigg|,
\end{equation}

where $D(q,\pointset)$ is the number of points of \pointset in the box $[0,q)$ and $\lebesguebox{q}$ is the Lebesgue measure of the $d$-dimensional box $[0,q)$.

Computing the \linf star discrepancy of a point set is a discrete problem as only points on a specific grid can reach the maximal value~\cite{niederreiter1972methods}. Intuitively, if one draws a grid using the coordinates of the points in a point set \pointset, then the maximum discrepancy can only be reached by boxes having their corner on that grid. Importantly, the evaluation of a point set in low dimensions can be done very efficiently with the DEM algorithm~\cite{dobkin1996computing} and its parallelizable implementation available from~\cite{clement2023computing}.

\subsection{Kronecker point sets}

There are various methods to generate point sets with low discrepancy. In 2 dimensions, the Fibonacci set provides good solutions. For a given $\nbpoint$, the Fibonacci point set $\fibopointset{\nbpoint}$ is obtained as follows:
\begin{equation}
    \fibopointset{\nbpoint}=\{(i/\nbpoint,i\phi\%1) \big| i \in \{0,\dots,\nbpoint-1\}\}
\end{equation}
where $\phi=(1+\sqrt{5})/2 \approx 1.618$ is the golden ratio and the notation $\%1$ means that we remove the integer part.

Kronecker sets are a generalization of Fibonacci. One can replace $\phi$ with other values and generate a point set in the same way. Given a set of parameters $\{p_2,\dots,p_d\}$, $p_j$ determining the coordinates in dimension $j$, one can then define a Kronecker point set as follows:

\begin{equation}
    \kroneckerpointset{\nbpoint}^{(p_2,\dots,p_d)}=\{(i/n\%1,i/p_2\%1,\dots,ip_d\%1)| \forall i \in\{0,\dots,n-1\}\}
\end{equation}

In 2 dimensions, one can slightly improve the performance of the Fibonacci point set by shifting the set and starting with $i=1$. We therefore define the shifted Kronecker set as follows.

\begin{equation}
    \shiftedkroneckerpointset{\nbpoint}^{(p_2,\dots,p_d)}=\{(i/n\%1,ip_2\%1,\dots,ip_d\%1)| \forall i \in\{1,\dots,n\}\}
\end{equation}

\begin{figure*}[h!]
    \centering
    \includegraphics[width=0.32\linewidth]{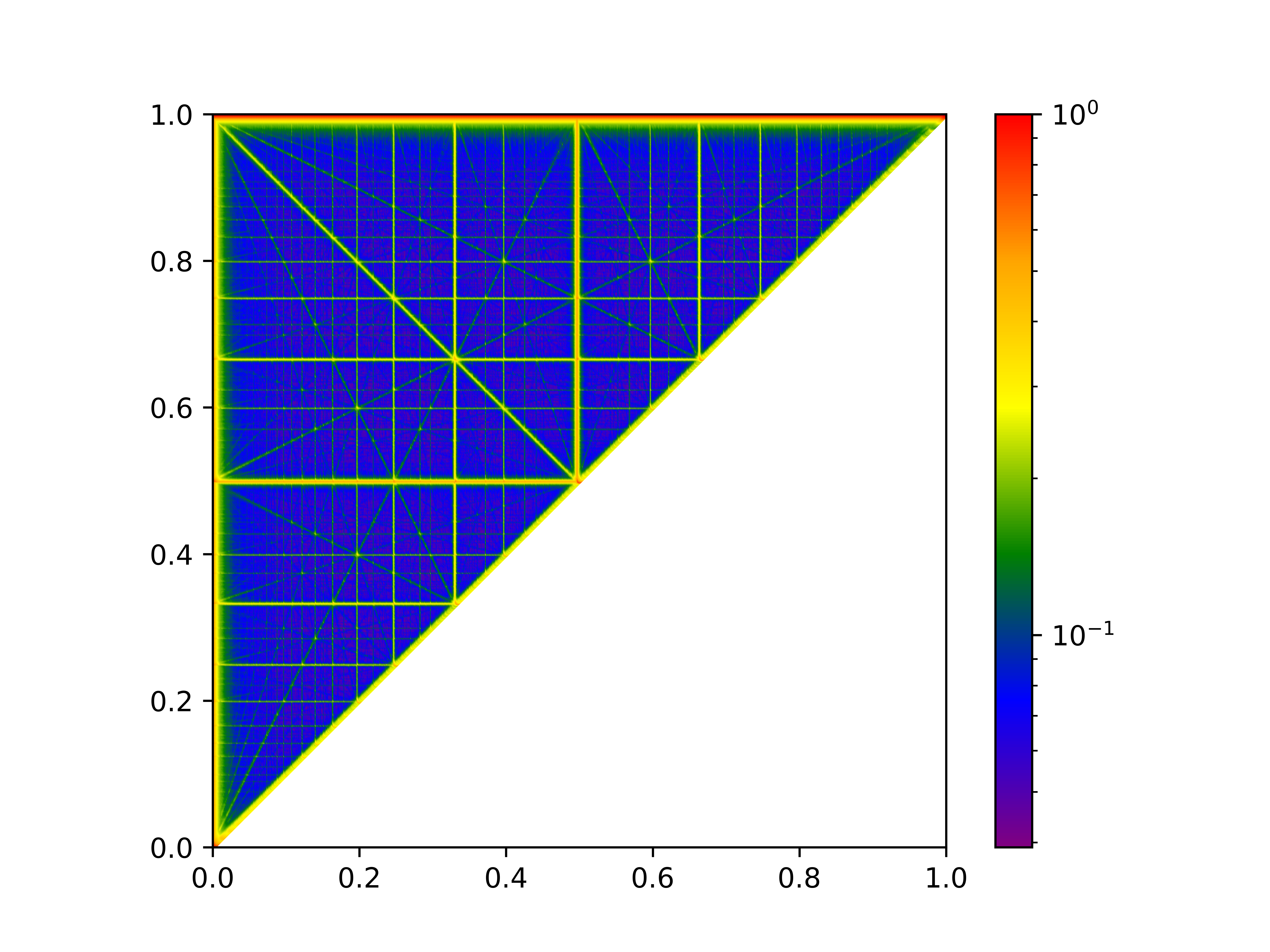}
    \hfill 
    \includegraphics[width=0.32\linewidth]{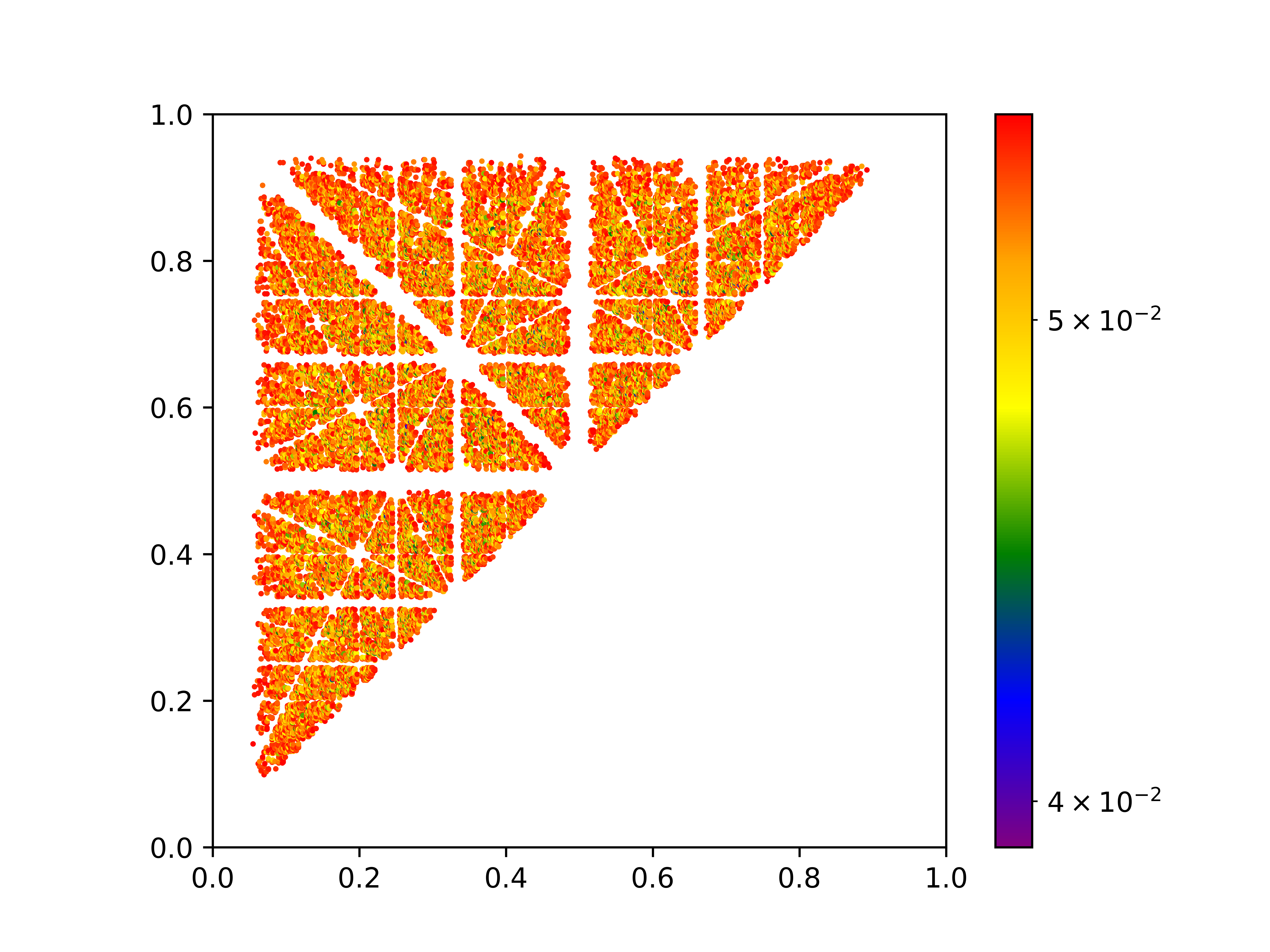} 
    \hfill
    \includegraphics[width=0.32\linewidth]{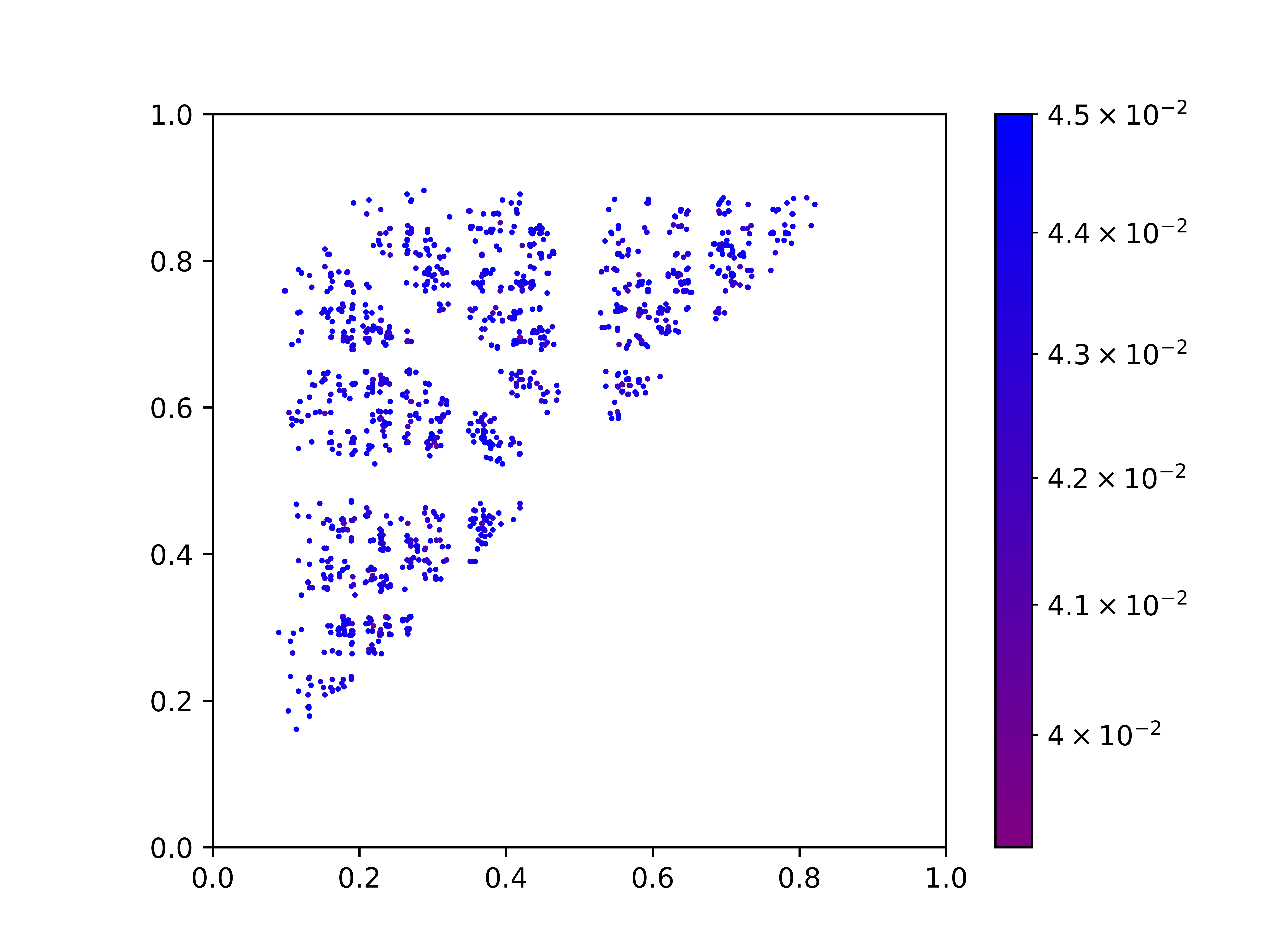}
    \caption{Heatmaps. The $X$ (resp. $Y$) coordinate corresponds to the $\parameterTwo$ (resp. $\parameterThree$) value, the color indicates the \linf star discrepancy of the 3D point set generated with the $(\parameterTwo,\parameterThree)\in [0, 1]^2$ parameters, for $\nbpoint=100$. The second and last heatmaps contain only points for which the corresponding Kronecker point set has a discrepancy below 0.055 and 0.045, respectively.}
    \label{fig:heatmap}
\end{figure*}

From now on, we use  ``Kronecker point set" instead of  ``shifted Kronecker point set" as experiments are only run for shifted sets. Additionally, since most of our experiments are conducted for point sets in 3 dimensions, we will focus on the 2-dimensional problem of finding values for parameters $p_2$ and $p_3$. 
Figure~\ref{fig:heatmap} displays heatmaps showing the quality of Kronecker point sets obtained with parameters $(\parameterTwo, \parameterThree)$ for $\nbpoint=100$. The $x$ and $y$ axis represent the parameters $p_2$ and $p_3$, respectively. Each dot on a heatmap represent the discrepancy of the point set generated with parameters $(p_2,p_3)$. On the left, we display all the points, in the center only points corresponding to point sets of discrepancy lower than $0.055$ and on the right only points corresponding to point sets with discrepancy lower than $0.045$. 

\begin{figure}[tb!]
    \centering
    \includegraphics[width=0.8\linewidth]{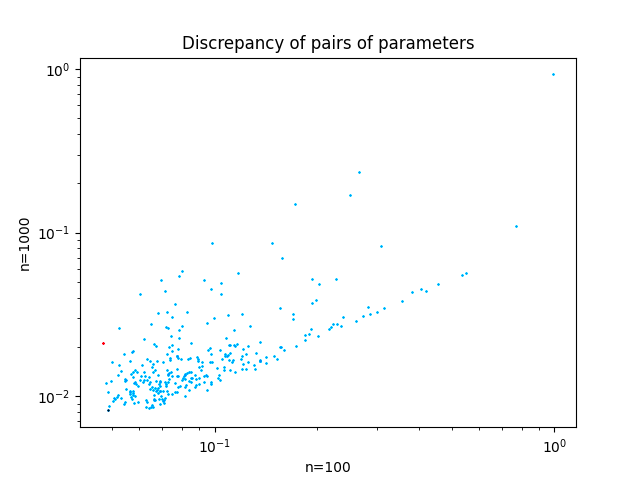}
    \caption{\linf star discrepancies of point sets obtained with randomly sampled $(p_2,p_3)$. The $X$ (resp. $Y$) coordinate of a point corresponds to the discrepancy of the point set obtained for $n=100$ (resp. $n=1000$). Red (resp. Black) point is the best found for $n=100$ (resp.$n=1000$).}
    \label{fig:disc100_1000}
\end{figure}

These heatmaps reveal that the optimization landscape is intricate to explore, as the problem appears highly multimodal. Very good values for parameters $\parameterTwo$ and $\parameterThree$ are scattered, and there seem to be many local optima. Additionally, this landscape depends on the value of \nbpoint and a good set of parameters for $\nbpoint=100$ is not guaranteed to be good for other values. Indeed, Figure~\ref{fig:disc100_1000} displays the discrepancies of the point sets obtained for $n=100$ and $n=1000$ using randomly sampled pairs of $(p_2,p_3)$.

Interestingly, while many very good pairs for $n=100$ seem to be also good for $n=1000$. There are also several pairs that provide a very low discrepancy point set for either $n=100$ or $n=1000$ but not both. In particular, one can see that the best pair found for $n=100$, represented by the red dot, does not yield a satisfactory point set for $n=1000$.

We are facing two challenges: 
(1) \textit{given a value \nbpoint, how can we find good parameter values $(p_2, p_3)$ that optimize the discrepancy of the set $\shiftedkroneckerpointset{\nbpoint}^{(1/\nbpoint, p_2, p_3)}$?}, (2) \textit{is it possible to find parameter values that return good point sets, for broad ranges of \nbpoint?}

\section{Optimizing Kronecker Point Set Generation}
\label{sec:algorithms}
In this section, we present the methods that we use to find Kronecker sets that minimize the discrepancy. We suggest using two optimization strategies: The first uses heuristics and black-box approaches to find specific values of the Kronecker parameters $(p_2,p_3) \in [0,1]^2$ for each number of points $\nbpoint$, where we consider $0 \leq p_2 \leq p_3 \leq 1$. 
Three heuristics are selected for the optimization of point generation: Covariance Matrix Adaptation Evolution Strategy (CMA-ES) \cite{hansen_cma-es}, Iterated Local Search (ILS) \cite{lourencco2018iterated} with snowball embedding local search, and Reactive Tabu Search (RTS) \cite{battiti1994reactive}, however, in lack of a clear difference in performance, we decide to continue only with CMA-ES for the later experiments for higher \nbpoint values, as this method is better established in the community.
The second strategy consists of applying the Automatic Algorithm Configuration (AAC) technique to tune Kronecker's parameters for points generation for selected intervals of $\nbpoint$ points, e.g., we find one configuration $(p_2,p_3) \in [0,1]^2$ 
that yields good point sets for all $\nbpoint \in [5,100]$. Algorithm configuration (AC) refers to the systematic process of tuning algorithm parameters and has been extensively studied in the literature ~\cite{stutzle2019automated,schede2022survey}. AAC, also called parameter tuning, aims to identify high-performing parameter settings without extensive manual intervention by framing parameter configuration as an optimization problem guided by empirical performance. This data-driven exploration improves robustness, reproducibility, and performance across diverse problem instances. Among state-of-the-art methods, \irace \cite{lopez2016irace} is widely adopted due to its statistically grounded racing mechanism, which efficiently focuses computational effort on the most promising configurations.

\section{Experimental Results}
\label{sec:experiments}

To evaluate the discrepancy values of our optimization techniques, we run several experiments to investigate the following research questions regarding the heuristics and automatic algorithm configuration.
\begin{itemize}
    \item (RQ1) How efficient are CMA-ES and parameter tuning for optimizing 3D Kronecker for star discrepancy?

    \item (RQ2) Is post-processing improving the Kronecker point sets performance?

    \item (RQ3) Using parameter tuning, can we find general $(p_2, p_3)$ pairs that are good for for ranges $[n_1,n_2]$ of \nbpoint?

    \item (RQ4) How efficient can the Kronecker method be if we increase the problem's dimension?
\end{itemize}

\subsection{Assessing the effectiveness of CMA-ES and parameter tuning for low discrepancy }

\begin{figure}[tb!]
    \centering
    \begin{subfigure}[t]{0.48\textwidth}      
        \includegraphics[width=7cm, height=5cm]{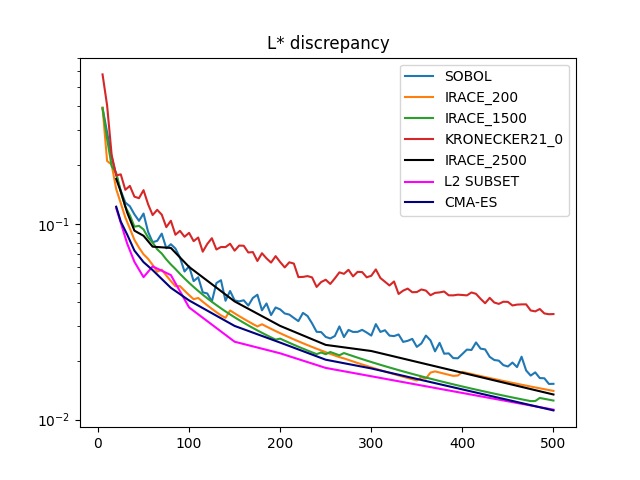}
        \subcaption{\nbpoint $\in [5, 500]$}
        \label{fig:plot_500}
    \end{subfigure}
    \quad
    \begin{subfigure}[t]{0.48\textwidth}
        \includegraphics[width=7cm, height=5cm]{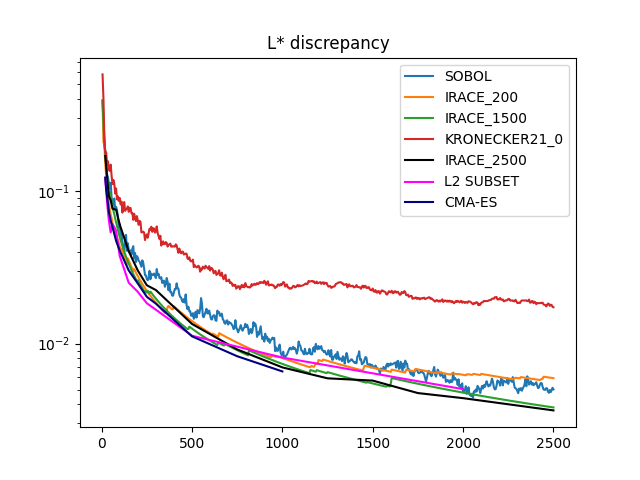}
        \subcaption{\nbpoint $\in [5, 2500]$}
    \label{fig:plot_2500}
    \end{subfigure}
    \caption{Plot of the \linf discrepancy for \nbpoint points}
\end{figure}

As the heuristic's computation time is expensive for a larger number of points, we limit experiments with CMA-ES to constructions of sets with at most \nbpoint = 1000 points. 
We run the algorithm five times with 10,000 evaluations per run. The reported result is the best found value among all evaluations. For the tuning of Kronecker's parameter settings, three scenarios were selected. In the first scenario, we tune \irace on a separate subselection for identifying good $(p_2, p_3)$ for ranges of n. 
Then, we compare our results in terms of quality of the returned discrepancy to the Sobol' sequence~\cite{Sobol}, for which we use the implementation by Fox \cite{fox1986algorithm}, and other state-of-the-art methods, namely KRONECKER21~\cite{patel2022optimizing}, L2\_Subset~\cite{clement2025low}. The results are reported in Table \ref{tab:Kronecker_3d} and Figures \ref{fig:plot_500} and \ref{fig:plot_2500}.

The parameter pair values ($p_2, p_3$)  obtained from the parameter tuning for I\_200, I\_1500, and I\_2500 are (0.5494, 0.7867), (0.6193, 0.7830), (0.71810558, 0.81422429), respectively.

\begin{table}[tb!]
\scriptsize
\caption{Best found \linf star discrepancy values for  $d = 3$}
\begin{center} 
\begin{tabular}{|c|c|c|c|c|c|c|c|}
\hline
n    & Sobol   & CMA-ES  & I\_200     & I\_1500   &I\_2500          & L2\_Subset          \\
\hline
20   & 0.17742  & 0.12221          & \cellcolor{black!10} 0.15029 & 0.18194          &  0.16935 & \textbf{0.1202}     \\
25   & 0.14742  & $\textbf{0.10066}^*$ & \cellcolor{black!10} 0.12738 & 0.14555          &   0.14618     & 0.1012         \\
32   & 0.14917 & 0.08813          & \cellcolor{black!10} 0.10094 & 0.12083          &    0.11420 & \textbf{0.07931} \\
40   & 0.11167  & 0.07292          & \cellcolor{black!10} 0.08273 & 0.09666          & 0.09221   &   \textbf{0.06392}  \\
50   & 0.11276  & 0.06371          & \cellcolor{black!10} 0.06986 & 0.09356          & 0.08682   &   \textbf{0.05337}  \\
60   & 0.08052  & \textbf{0.05799} & \cellcolor{black!10} 0.06166 & 0.08021          & 0.07639   & 0.0606             \\
80   & 0.07862 &  \textbf{0.04709} & \cellcolor{black!10} 0.05099 & 0.06199          & 0.07525  &    0.0547              \\
100  & 0.06057  & 0.04017          &  \cellcolor{black!10} 0.04325 & 0.04996          & 0.06020   & \textbf{0.0374}         \\
150  & 0.04062 & 0.03009          & 0.03489 & \cellcolor{black!10} 0.03331          &    0.04014 &    \textbf{0.02499}     \\
200 & 0.03654   &   0.02474 &   0.02762 &   \cellcolor{black!10} 0.02592 &   0.03010 &   \textbf{0.02181} \\
250  & 0.02645 &  0.02023          & 0.02210 & \cellcolor{black!10} 0.02158          &  0.02408  & \textbf{0.01837}    \\
300  & 0.02686 &  \textbf{0.01826} & \cellcolor{black!10} 0.01849 & 0.01969          &     0.02241   &                 \\
500  & 0.01524 &  \textbf{0.01115} & 0.01402 & \cellcolor{black!10} 0.01252     & 0.01344     & 0.01125               \\
750  & 0.01236 &  \textbf{0.00826} & 0.01014 & \cellcolor{black!10} 0.00904          &      0.00915 &                   \\
1000 & 0.00836 &       \textbf{0.00657}           & 0.00804 & 0.00736    &   \cellcolor{black!10} 0.00698    & 0.0081              \\
1250 & 0.00853 &                  & 0.00770 & 0.00650 &   \cellcolor{black!10} \textbf{0.00592} &                   \\
1500 & 0.00713 &                 & 0.00695 & \cellcolor{black!10} \textbf{0.00548} &     0.00572 &               \\
1750 & 0.00649 &                  & 0.00677 & 0.00545 &    \cellcolor{black!10} \textbf{0.00474} &                   \\
2000 & 0.00480 &                  & 0.00624 & 0.00477 &   \cellcolor{black!10} \textbf{0.00439} & 0.00503                \\
2500 & 0.00501 &                  & 0.00593 & 0.00382 &  \cellcolor{black!10} \textbf{0.00365}   &
\\
3000 & 0.00435 & & 0.00537 & 0.00328 & \cellcolor{black!10} \textbf{0.00304} &  \\
3500 & 0.00383 & & 0.00520 & 0.00314 & \cellcolor{black!10} \textbf{0.00268} &  \\
4000 & 0.00314 & & 0.00501 & 0.00286 & \cellcolor{black!10} \textbf{0.00262} &  \\
5000 & 0.00264 & & 0.00485 & 0.00241 & \cellcolor{black!10} \textbf{0.00210} &  \\
\hline
\end{tabular}
\end{center}
\begin{tablenotes}
      \item {\small{
      \textbf{Note:} The value with $^*$ is the smallest value found by any of the three evolutionary heuristics. We highlight in gray the cell containing the best value among the three \irace experiments.}}
\end{tablenotes}
\label{tab:Kronecker_3d}
\end{table}

Table \ref{tab:Kronecker_3d} displays \linf star discrepancy values for chosen \nbpoint points in dimension $D = 3$. The results show that our CMA-ES is competitive for small \nbpoint and sometimes better than the state of the art. For sizes 25, 60, 80, and \nbpoint $>$ 300, we outperform L2\_subset and found new low discrepancy values. Additionally, for higher values of \nbpoint $\geq 500$, CMA-ES and the parameter tuning I\_1500 and I\_2500 using \irace have outperformed the L2\_subset and Sobol and achieved new state-of-the-art discrepancy values.

Furthermore, we compare the values returned by \irace in the different experiments. In Table~\ref{tab:Kronecker_3d} we highlight cells with the lowest discrepancy value among the \irace experiments. The tuning is generally adapted to the size of \nbpoint on which it is tuned, but gets slightly worse for bigger \nbpoint values that were not seen in the training. 
This is particularly relevant as one can see in Figure~\ref{fig:plot_2500} that the parameters obtained with I\_200 seem to produce significantly worse point sets for higher values of \nbpoint. The opposite does not seem to hold as I\_1500 and I\_2500 produce acceptable point sets for smaller sizes, although they were trained to also fit much larger values of \nbpoint. 
Finally, one can see that I\_2500 keeps outperforming the Sobol sequences even outside of its training space for sizes 3000, 3500, 4000, and 5000.

\begin{figure}[tb!]
    \centering
    \includegraphics[width=0.72\linewidth]{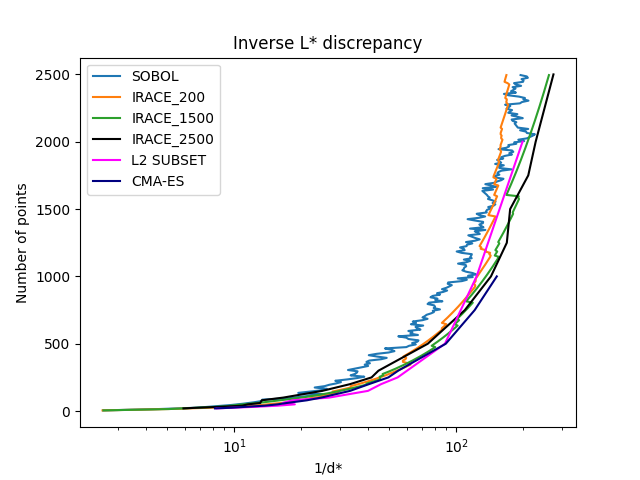}
    \caption{Plot of the inverse \linf star discrepancy for \nbpoint between 5 and 2500}
    \label{fig:plot_reversed}
\end{figure}

Figure \ref{fig:plot_reversed} displays the inverse \linf star discrepancy, i.e., the number of points required to achieve a target discrepancy. For readability, the $x$ axis shows the inverse of the discrepancy. This figure can be read as follows: "Given an objective on the discrepancy, how many points are needed?". Overall, the results show that our optimized approaches outperform classical constructions. In particular, the Sobol sequence exhibits higher variability and generally requires more points, highlighting its limitations in finite-sample regimes despite strong asymptotic guarantees. The L2\_Subset method performs competitively for moderate discrepancy levels but scales less effectively as stricter targets are imposed. In contrast, CMA-ES and \irace-based configurations consistently achieve lower sample requirements across a wide range of discrepancy values. CMA-ES is particularly effective in the low-to-medium regime, while \irace demonstrates superior scalability and robustness, especially for tight discrepancy thresholds. These results emphasize the advantage of adaptive, black-box optimization methods over static constructions, enabling more efficient generation of low-discrepancy point sets for practical applications. Finally, this shows a great improvement if it were applied for real-world applications that require big sample for their experiments, where with CMA-ES and \irace-based results, we can minimize the size of the sample to achieve a certain target discrepancy, e.g;, for an inverse discrepancy of $10^2$, with the L2\_Subset method, we need a sample of size of about 700 points to reach this discrepancy value, where with CMA-ES we need only 500 points, and this difference can result in a big gain of time process for the real world application.

\vspace{-0.5cm}

\subsection{Using post-processing on Kronecker point sets} 
Clément et al.~\cite{clement2025searching} propose a method to improve the discrepancy starting from any given point set. Their method maintains the relative position of the points, i.e., for two points $\point{i}$ and $\point{j}$, if \point{i} has a smaller coordinate in any given dimension $k$ than \point{j}, then this will still be the case after the post-processing step. It returns the best possible point set such that this relative position is maintained using mathematical programming. This post-processing is, however, rather costly and therefore can only be applied for small values of $\nbpoint$. Results can be found in~ Table~\ref{tab:post-processing}.

\begin{table}[h!]
\scriptsize
    \centering
    \caption{\linf star discrepancy with post-processing. Notation o(name) means after post-processing.}
    \begin{tabular}{|c|c|c|c|c|c|c|}
    \hline
        \nbpoint & I\_200 & o(I\_200) & L2\_Subset & o(L2\_Subset) & MPMC & o(MPMC)\\
        \hline
         25 & 0.12738 & 0.11029 & 0.1012 & 0.08519 & 0.10664 & \textbf{0.08335}\\
         32 & 0.10094 & 0.08956 & 0.07931 & 0.07309 & 0.08234 & \textbf{0.07085}\\
         40 & 0.08273 & 0.06082 & 0.06392 & \textbf{0.05988} & 0.08139 & 0.06242\\
         50 & 0.06986 & 0.05420 & 0.05337 & \textbf{0.04979} & 0.05828 & 0.05067\\
         \hline
    \end{tabular}
    \label{tab:post-processing}
\end{table}

Unsurprisingly, point sets that are better before post-processing remain better afterwards. However, one can note that the Kronecker point sets seem to benefit more from the post-processing. This may be due to the fact that coordinates for Kronecker point sets are generated in a very rigid and generic way and are more likely to benefit from small modifications than point sets obtained through optimization techniques.

\subsection{Assessing Parameter tuning on different \nbpoint ranges}
\begin{figure}[tb!!]
    \centering
    \includegraphics[width=0.72\linewidth]{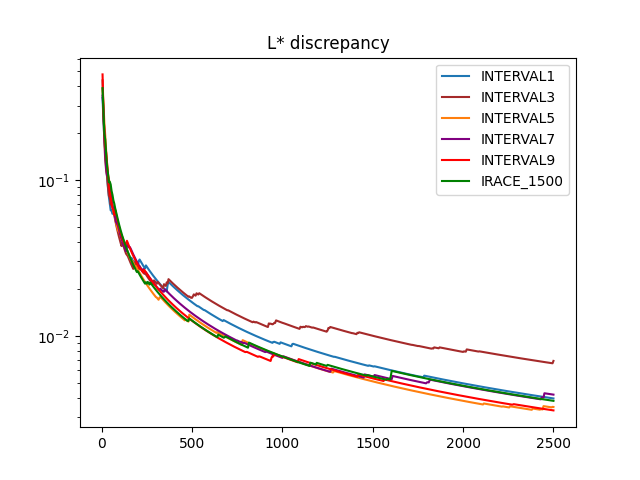}
    \caption{$\linf$ discrepancy ($y$-axis) with Parameters tuning for \nbpoint small intervals (problem size on $x$-axis)} 
    \label{fig:irace_chunck_complete}
\end{figure}

We compare the evolution of the $\linf $ discrepancy for multiple interval configurations  (INTERVAL1 - INTERVAL10) of $\nbpoint \in [5,1000]$ and the obtained configuration \irace\_1500 in Table \ref{tab:Kronecker_3d}. INTERVAL1 means $\nbpoint \in [5, 100]$, INTERVAL2 has $\nbpoint \in [101, 200]$ and so on until INTERVAL10 that has $\nbpoint \in [901, 1000]$. The main objective of this study is to find good pairs $(\parameterTwo, \parameterThree) \in [0,1]^2$ that are good for a specific interval but that could also be good as well for all \nbpoint. Then we rerun the algorithm for the whole \nbpoint $\in [5, 2500]$. To have a clear vision on the results, we keep only the results for odd-number intervals as shown in the plots in Figure~\ref{fig:irace_chunck_complete}.

Figure \ref{fig:irace_chunck_complete} illustrates how the lines are closely packed when \nbpoint is small, suggesting that most configurations perform similarly for smaller sizes. Performance differences become more noticeable as the number of points increases. For example, INTERVAL\_5 and INTERVAL\_9 consistently outperform the majority of other interval configurations and are competitive with \irace\_1500 up to about 1000–1200 points; beyond this range, a clear improvement emerges. 
We do not have a full understanding of why this happens. This suggests that there exists some space to further improve the Kronecker sets beyond the results we already obtained with \irace. 

However, in addition to demonstrating the benefit of data-driven parameter tuning in navigating the highly nonconvex discrepancy landscape and preventing premature performance saturation, this suggests that automatic algorithm configuration is more effective at identifying parameter settings that generalize across increasing sample sizes.

\subsection{Assessing the Kronecker method for dimension $D = 4$}
For further study, we evaluate the Kronecker method and the optimized approaches for dimension $D = 4$. This is the only section for which we optimize three parameters $(p_2,p_3,p_4)$ instead of just $(p_2,p_3)$. Results are reported in Table~\ref{tab:kronecker_4d}. We observe that even the CMA-ES and RTS algorithms that tune the parameters specifically for one given \nbpoint cannot outperform the discrepancy values of the truncated Sobol' sequence in 4 dimensions. One question that remains is whether this is due to the nature of Kronecker sets or to the ability of the algorithms to find very good parameters. The former would mean that Kronecker sets would be unable to provide good point sets for dimension~$D=4$, and possibly beyond. 
\begin{table}[t]
\caption{\linf discrepancy values of the best-found Kronecker constructions in dimension 4, for different values of $n$ (3 columns on the right), compared to those obtained from the Sobol' sequence\\}
\centering
\begin{tabular}{|l|l|l|l|l|}
\hline
\multicolumn{1}{|c|}{$n$} & \multicolumn{1}{c|}{Sobol'} & CMA-ES & \multicolumn{1}{c|}{RTS} & \multicolumn{1}{c|}{\irace\_512} \\
\hline
5                     & 0.28281                   &  0.34054   & 0.34184                 & 0.43494263                     \\
8                     & 0.23438                   &  0.29388   & 0.27840                 & 0.32895087                     \\
16                    & 0.13672                   &  0.17685   & 0.19424                 & 0.24601968                     \\
20                    & 0.11172                   &  0.15750   & 0.16216                 & 0.22276414                     \\
32                    & 0.08984                   &  0.12989   & 0.12888                 & 0.1486705                      \\
50                    & 0.07994                   &   0.08886  & 0.09975                 & 0.09884006                     \\
64                    & 0.05371                   &   0.07782  & 0.08338                 & 0.10571434    \\
\hline
\end{tabular}
\label{tab:kronecker_4d}
\end{table}

\section{Analysis of the Problem Landscape}
\label{sec:landscapeAnalysis}

\begin{figure}[h!]
    \centering
    \begin{subfigure}[t]{0.45\textwidth}
        
            \includegraphics[width=6.3cm, height=4cm]{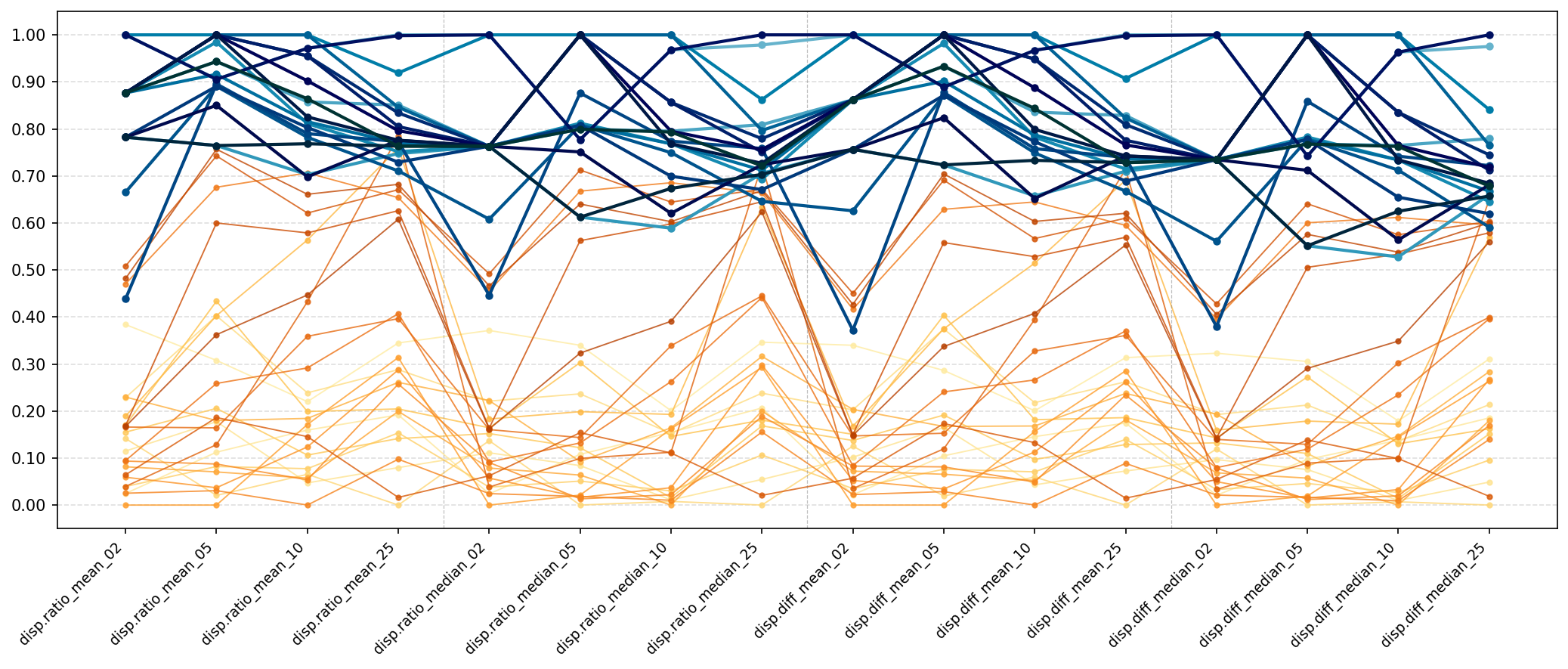}
            \subcaption{dispersion features}
            \label{fig:dispersion_ft}
    \end{subfigure}
\quad
    \begin{subfigure}[t]{0.45\textwidth}
        
        \includegraphics[width=6.3cm, height=4cm]{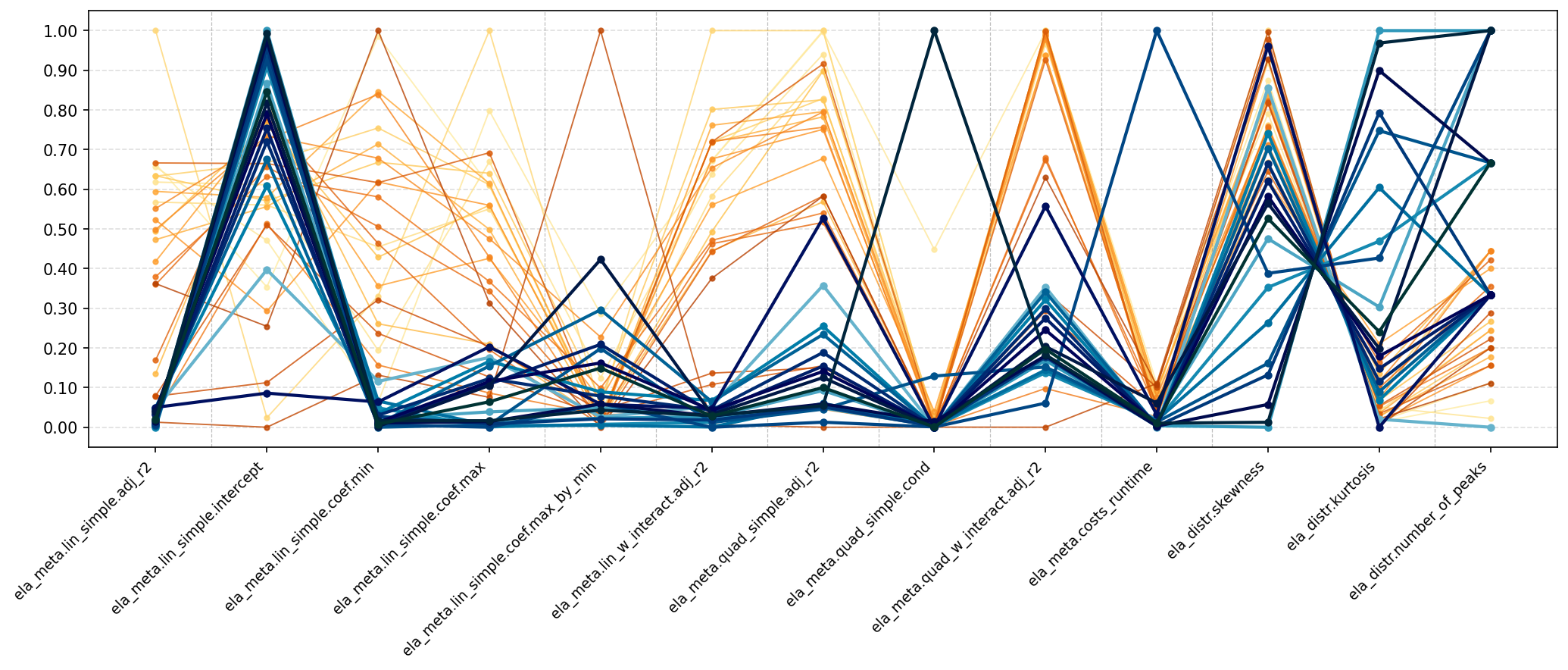}
        \subcaption{ela features}
        \label{fig:ela_ft}    
    \end{subfigure}
    \quad
    \begin{subfigure}[t]{1\textwidth}
        \centering
        \includegraphics[width=8.3cm, height=5cm]{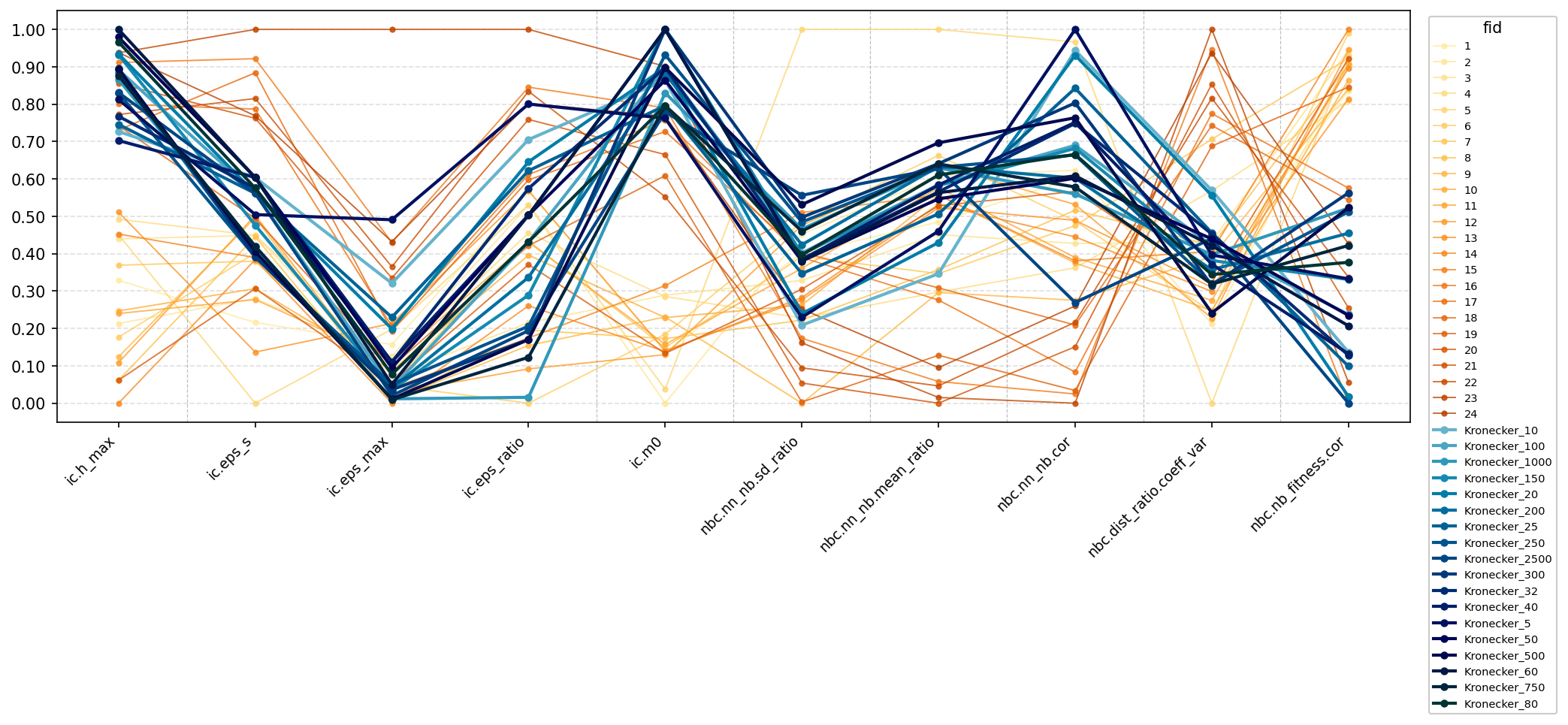}
        \subcaption{information content, and nearest better clustering features}
        \label{fig:cellmapping_ft}    
    \end{subfigure}
    \caption{Comparison of selected feature values of the 3D kronecker problem (blue lines) with those of the 24 BBOB functions (slim red lines). All values have been normalized to the interval [0, 1]}
    \label{fig:features_analysis}
\end{figure}

We have already observed in Figure \ref{fig:heatmap} that the optimization landscape is highly multimodal. To further probe the characteristics of the optimization problem, we investigate the Kronecker configuration problem using exploratory landscape analysis (ELA) \cite{mersmann2011exploratory_ela}. We compare the ELA features using the pflacco package \cite{10.1162/evco_a_00341_pflacco} with its default setting, which samples $50 \times D = 100$ points uniformly at 'LHS'. We evaluate 18 samples and compare the averages of the feature values of 15 evaluations with those of the 24 BBOB functions of the COCO framework \cite{Hansen02012021_coco}, computed using the exact same protocol. 
The results of the comparison are illustrated in Figure \ref{fig:features_analysis}. We see that the 3D Kronecker problem has high dispersion feature values compared to all the 24 BBOB functions (see Figure \ref{fig:dispersion_ft}), additionally, we see higher values for the distribution kurtosis feature, and the number of peaks in the ELA features (Figure \ref{fig:ela_ft}), which gives insights of the high multimodality landscape of the problem, and for the other feature sets displayed, we see that no BBOB function could be identified for a good fit in Figure \ref{fig:ela_ft} and \ref{fig:cellmapping_ft}. Additionally, we can also see from the figures that the landscape for the Kronecker problem is closely similar for the different instances of the same problem. 

\section{Discussion}
\label{sec:discussion}
Using optimization strategies based on the CMA-ES heuristic and automatic algorithm configuration, we have optimized the parameters of Kronecker sets for the generation of sets of low star discrepancy. Recent state-of-the-art approaches, such as the L2\_Subset method \cite{clement2025searching}, are particularly effective for constructing low-discrepancy point sets with a small number of points. Our results show that the CMA-ES heuristic is similarly competitive in the small to medium regime of point-set sizes \nbpoint compared to L2\_subset. In addition, automatic algorithm configuration using \irace proves more effective for larger point sets, where systematic exploration of the parameter space yields superior performance. 
However, while we obtain good results in 3 dimensions, the Kronecker set seems to perform worse in 4 dimensions. 

\begin{figure}[tb!]
    \centering
    \includegraphics[width=0.65\linewidth]{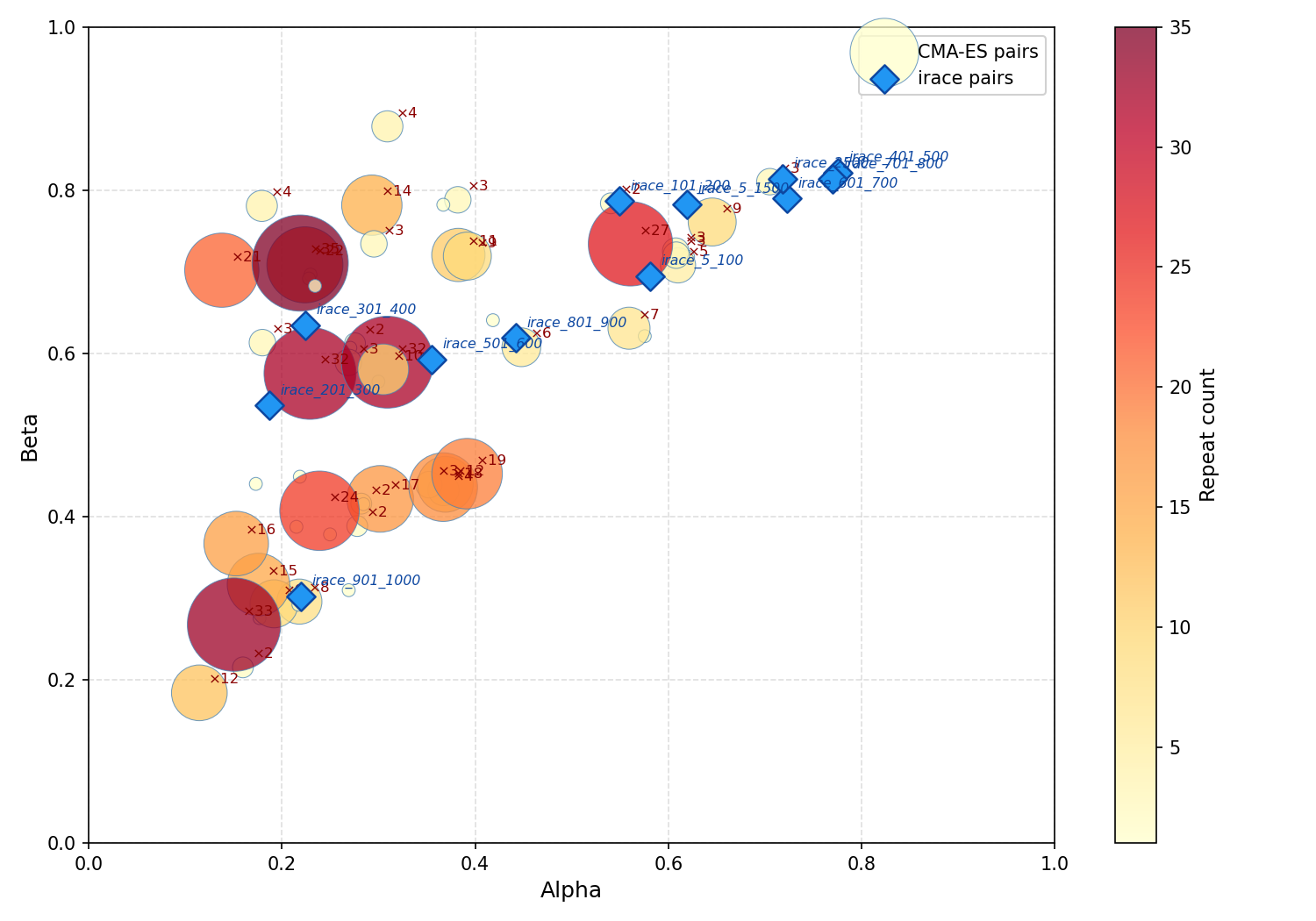}
    \caption{Distribution of ($\alpha, \beta$) parameter pairs}
    \label{fig:apha_beta_values}
\end{figure}

Figure \ref{fig:apha_beta_values} shows the distribution of ($\alpha, \beta$) parameter pairs obtained by CMA-ES and \irace. CMA-ES explores a wide range of configurations, with dense clusters around $\alpha \in [0.2,0.4]$ and varying $\beta$, indicating convergence to multiple local optima and highlighting the multimodal nature of the problem. Frequently repeated points suggest the presence of several competitive parameter settings. In contrast, \irace exhibits a more structured trajectory, progressively favoring larger $\alpha$ and $\beta$ values as the problem size increases. While both methods overlap in some robust regions (e.g., $\alpha \approx 0.25$, $\beta \approx 0.6$), \irace identifies additional high-value configurations that CMA-ES does not consistently reach.

Overall, this study highlights the complexity of the parameter optimization problem and underscores the value of combining different optimization paradigms.

\section{Conclusion}
\label{sec:conclusion}

We have used algorithm configuration techniques to find good parameters for Kronecker sets in 3 dimensions. The point sets generated are competitive with state-of-the-art methods for small set sizes and consistently outperform the previously best known constructions for point sets containing at least 300 points.

Our experiments reveal that there may be some space for further improvements when using Kronecker sets in dimension $D=3$. In dimension $D=4$ (and probably beyond), however, Kronecker sets do not seem to be able to outperform standard constructions like truncated Sobol' sequences. 

Our results call for a better understanding of the parameter to discrepancy mapping for Kronecker sets in $D=3$. We believe that further improvements might be possible. We also believe that a finer understanding of the trade-off between size-specific constructions and parameters that yield good sets across broader ranges of set sizes would be of great practical relevance, as in real-world situations, it may not be uncommon that an existing design of experiment or an initial sample can be extended by additional points.  

Finally, our results also indicate a lack of high-performing constructions for dimensions $D\ge4$. Here, the subset selection approaches proposed in~\cite{clement2024heuristic,ClementDP22} seem to be state of the art. However, with these being dependent on the sets that the subsets are taken from, we expect significant potential in the optimization of low-discrepancy point sets for these settings.

Last but not least, from an evolutionary computation perspective, we have demonstrated that the construction of low-discrepancy point sets offers challenging optimization problems of high practical relevance. We expect to see further contributions of our community to this important application.

\vspace{2ex}
\textbf{Acknowledgments.} 
This research was supported in part by the French PEPR integrated project HQI (ANR-22-PNCQ-0002), by the French National Research Agency (ANR-23-CE23-0035) and the German Research Foundation (DFG; LI 2801/7-1), through project \textsc{Opt4DAC}, and by the European Union (ERC, ``dynaBBO'', grant no.~101125586). Views and opinions expressed are, however, those of the author(s) only and do not necessarily reflect those of the European Union or the European Research Council Executive Agency. Neither the European Union nor the granting authority can be held responsible for them.

%
%
%
\bibliographystyle{splncs04}
\bibliography{Mybib}

\end{document}